\documentclass[a4paper,twoside]{article}

\pdfoutput=1

\usepackage{epsfig}
\usepackage{subcaption}
\usepackage{calc}
\usepackage{amssymb}
\usepackage{amstext}
\usepackage{amsmath}
\usepackage{amsthm}
\usepackage{multicol}
\usepackage{pslatex}
\usepackage{apalike}
\usepackage{SCITEPRESS}     

\usepackage[nolist,nohyperlinks]{acronym}

\begin{acronym}
\acro{3D}{Three-Dimensional}
\acro{CNN}{Convolutional Neural Network}
\acro{HCI}{Human Computer Interaction}
\acro{VR}{Virtual Reality}
\acro{RGB}{Red Green Blue}
\acro{JSON}{JavaScript Object Notation}
\acro{GloVe}{Global Vectors for Word Representation}
\acro{ZSL}{Zero-shot learning}
\acro{MRF}{Markov Random Field}
\acro{DCNN}{Deep Convolutional Neural Network}
\acro{CRF}{Conditional Random Field}
\acro{RGBD}{Red Green Blue - Depth}
\acro{ReLU}{Rectified Linear Unit}
\acro{RWA}{Radially Weighted Average}
\acro{LRN}{Local Response Normalization}
\acro{CSV}{Comma Separated Values}
\acro{POV}{Point of View}
\acro{YAML}{YAML Ain't Markup Language}
\end{acronym}

\begin{document}

\title{Predicting Relative Depth  between Objects from Semantic Features}

\author{\authorname{Stefan Cassar\sup{1}
, Dylan Seychell\sup{1}\orcidAuthor{0000-0002-2377-9833} and Adrian Muscat\sup{2}\orcidAuthor{0000-0002-9157-2818}}
\affiliation{\sup{1}Department of Artificial Intelligence, University of Malta, Msida, Malta}
\affiliation{\sup{2}Department of Computer Engineering, University of Malta, Msida, Malta}
\email{\{stefan.j.cassar.13, dylan.seychell, adrian.muscat\}@um.edu.mt}
}

\keywords{Relative Depth, Monocular Image, Computer Vision, Image Semantics}


\abstract{Vision and language tasks such as Visual Relation Detection and Visual Question Answering benefit from semantic features that afford proper grounding of language.  The 3D depth of objects depicted in 2D  images is one such feature. However it is very difficult to obtain accurate depth information without learning the appropriate features, which are scene dependent. The state of the art in this area are complex Neural Network models trained on stereo image data to predict depth per pixel.  Fortunately, in  some tasks, its only the relative depth between objects that is required.  In this paper the extent to which semantic features can predict course relative depth is investigated.  The problem is casted as a classification one and geometrical features based on object bounding boxes, object labels and scene attributes are computed and used as inputs to pattern recognition models to predict relative depth. i.e behind, in-front and neutral. The results are compared to those obtained from averaging the output of the \textit{monodepth} neural network model, which represents the state-of-the-art. An overall increase of 14\%  in relative depth accuracy over relative depth computed from the \textit{monodepth} model derived results is achieved.
}

\onecolumn \maketitle \normalsize \setcounter{footnote}{0} \vfill

\section{Introduction}

Depth estimation is the result of a process which yields depth information of pixels, segments or objects in an image. Extracting information about the depth of an object is a challenging problem yet plays an important role in various computer vision tasks, including semantic labelling \cite{Ladicky_2014_CVPR}, robotics \cite{Hadsell2009}, pose estimation \cite{Shotton2013}, recognition \cite{ren2012rgb}, \ac{HCI} \cite{fanello2014learning}, \ac{VR} \cite{Li2017} and scene modelling \cite{Hoiem2005,Saxena2009}. Given the wide use of depth annotations of an object in an image in several computer vision applications, tackling the problem of obtaining depth from a monocular image is of vital use and importance. Vision and language tasks such as Visual Relation Detection benefits from depth features that afford proper grounding of language, \cite{birmingham2018adding}.

However, the problem of depth estimation of a monocular image poses some problems since it is difficult to obtain a good measurement of depth by using a single image without extracting any feature information from it or without matching the scene of the image with a \ac{3D} one \cite{Li2015}. A coloured image of a particular scene may have numerous \ac{3D} representations that can be used to obtain its depth \cite{Afifi2016}. Thus, estimating depth by attempting to match the image to existing \ac{3D} scenes composed of the same objects can prove to be inaccurate. For this reason, the importance of extracting ``monocular depth cues such as line angles and perspective, object sizes, image position and atmospheric effects'' for depth estimation on single images, is further emphasised \cite{Eigen}.

This work aims to make use of semantic information extracted using computer vision techniques such as object detection, scene detection, perceptual and quantitative spatial relations in between  objects to detect the relative depth between two objects in a monocular image using a supervised classification approach.  The term relative depth refers to the course depth of an object relative to another. In classification terms relative depth takes one of three categories; \textit{in front of}, \textit{behind} or \textit{neutral}. To train the model, a dataset composed of \ac{RGB} images labelled with objects with known depth information along with other descriptive cues and features is used. Different combinations of image features are also used to assess the usefulness and weight of specific scene cues when constructing the depth estimation model. 
The results obtained by the models in this work are compared with the results derived from the depth maps obtained from the state-of-the-art  \textit{monodepth} predictor. 
The rest of the paper is organised as follows. Section 2 describes related work; section 3 the methodology, including dataset, features and models used;  section 4 describes the results; section 5 evaluates the model against results derived from the \textit{monodepth} model; and section 6 concludes the paper.

\section{Related Work}

The problem of depth estimation has been tackled in past literature using techniques such as structure from motion \cite{crandall2011discrete}, n-view reconstruction \cite{wu2011repetition} and also simultaneous localisation and mapping \cite{salas2013slam}. However, tackling the task of estimating depth using a monocular image has not been as widely explored. A monocular image may be considered as ambiguous since it does not contain any depth information and may be represented in various \ac{3D} representations. Scene specific knowledge may aid the depth estimation process by making assumptions using the geometric properties of the scene \cite{Li2015}. However, the relevance of this information is limited to how close the image represents a specific structure.

\cite{saxena2006learning}, use a \ac{3D} scanner to collect images and their corresponding depth maps using a SICK 1-D laser range finder. This dataset is composed of 425 images and depth pairs of outdoor and indoor places. They segment the image into equal sized patches and calculate a single depth value for each patch. In their work, texture variations, texture gradients and haze are features in an image that are used to calculate depth. The researchers differentiate between absolute features, which for a patch are used to approximate the absolute depth. They also consider relative features to estimate the depth relative to the change in depth between the two patches \cite{saxena2006learning}. Texture information is extracted from the intensity channel using Laws’ masks. For depth estimation of a patch, \cite{saxena2006learning} argue that features relative to the selected patch are insufficient and therefore consider global properties of the image at multiple scales since at different depths (and therefore scales), objects have different properties, and thus any variations in the scales can be modelled using these different scale instances. \cite{saxena2006learning} report positive results produced by their depth estimation model, yet point out that the model struggles to produce correct absolute depth values when ``irregularly shaped trees'' are present in the image \cite{saxena2006learning}.

\cite{wu2011repetition} extract the repetition and symmetry of patterns and use this information to model the reconstruction of the shapes in the image. They enforce constraints using a penalisation scheme on inconsistencies between the repetition of pairs of pixels \cite{wu2011repetition}. \cite{fouhey2014unfolding} propose the use of edges in convex or concave form to interpret the scene and produce a \ac{3D} representation of the input image \cite{fouhey2014unfolding}. In earlier work, \cite{zhang1999shape} attempt to deduce an object by processing the shading information of the object in the image. They implement six shape-from-shading algorithms and conclude that an ideal case does not always hold for all input images, since these geographic attributes may not always be available or cannot be inferred from the image \cite{Li2015}. \cite{Saxena2009} also attempted to create accurate \ac{3D} models from an image by using supervised \ac{MRF}s in which an image is segmented using over-segmentation into about 2,000 superpixels. Superpixels are regions that are similar and were used as information for the depth calculation process. Contrary to other work that makes use of shape from shading \cite{zhang1999shape,maki2002geotensity} or shape from texture \cite{malik1997computing,lindeberg1993shape}, the researchers developed a probabilistic model which makes use of \ac{MRF}s to extract information of a region in relation to another region. Subsequently, each superpixel is enhanced with features that describe the area in the superpixel. These cues allow for the construction of meaningful \ac{3D} scenes. Using object detection and identification techniques, \cite{Saxena2009} enhance the \ac{MRF} with constraints using objects that appear in the scene that validate the relationship between objects such as ``on top of'' and ``attached to''. This method produced \ac{3D} models which are 64.9\% accurate without assuming any structural features of the scene \cite{Saxena2009}.

\cite{Li2015} apply deep learning using a multi-stage combinatory approach that makes use of \ac{DCNN} and \ac{CRF} for pixel-level depth estimation. They refine superpixels to pixel level rather than the other way round. Superpixels are acquired to represent different scale patches around the origin point of the superpixel center. A deep \ac{CNN} is used to produce a relationship between the input superpixel and the depth. The depth value is refined to pixel level from super pixel level using a \ac{CRF} without using geometric information assumed from the image scene. Furthermore, cues such as occlusion are inferred from the large amounts of training data, thus displaying a high level of generalisation \cite{Li2015}. 

In their work, \cite{Liu2014} modelled the problem as a “discrete-continuous optimisation problem” and used \ac{CRF}s to provide depth values by encoding superpixels and the relationships between them. For an input image, similar images were retrieved from a dataset of images with depth information. Each superpixel was rewritten as a value that describes the depth between the centroid and its plane. Similar to the work of \cite{Saxena2009}, \cite{Liu2014} make use of the relationship between two superpixels. They use discrete variables to encode the relationship between adjacent pixels. \cite{Liu2014} report positive results on the \ac{3D} models generated \cite{Liu2014}.

\cite{Cao2017} applied a deep network to obtain the depth information for each pixel in an image. A classification approach is followed in which multiple labelled categories were formed by taking into consideration different ranges of the depth values. This approach is contrary to most of the other approaches that tackle the problem as a regression problem. The deep network was then applied to label each pixel with an estimated depth. The continuous values obtained are separated into several discrete bins. As a post-processing step, the results obtained were then enhanced by applying a fully connected \ac{CRF}. This allows for the improvement of pixel level depth estimation by using other pixels that are connected to it \cite{Cao2017}.

\cite{Liu2010} couple the problem of depth estimation with that of semantic segmentation. They base the depth estimation process on the semantic information obtained from scene understanding. This approach is composed of two stages: (1) prediction of the semantic class for the pixels of the image through the use of ``a learned multi-class image labelling \ac{MRF}''. Each pixel was labelled as either ``sky, tree, road, grass, water, building, mountain or foreground object''. These labels were carefully selected to cater for backgrounds in outdoor scenes, as well as for any foreground objects in scenes. (2) The labels were then used as inputs to the depth estimation model. For each semantic class, a separate depth estimator was implemented. Each semantic label has scene specific information encoded in the \ac{MRF}. \cite{Liu2010}'s solution is a continuation of the work of \cite{Saxena2009} \cite{Saxena2009} and involves the use of pixel-based and superpixel-based variations of the constructed model. Their work was validated against \cite{Saxena2009}'s work as well as other state of the art methods and produces better scene reconstructions \cite{Liu2010}.

\cite{zeisl2014discriminatively} also make use of contextual cues. Without assuming any cues from the \ac{3D} scene, \cite{zeisl2014discriminatively} utilise previous work conducted in image labelling tasks and assume that the pixels located in a detected segment will contain the same label. The authors note that this notion holds for surfaces that are flat. For curved edges, the cues extracted at pixel level are combined with those extracted using detectors for segmentation cues. The input images are segmented to extract the dense and discriminating features. When testing, the classifier is used to predict the probability of the ``representative normals''. \cite{zeisl2014discriminatively} report positive results obtained using the classifier which ``successfully recognises surface normals for a wide range of different scenes'' \cite{zeisl2014discriminatively}.

In their work, \cite{lin2019depth} presented a hybrid convolutional neural network that incorporates together semantic segmentation and depth estimation. They first estimate depth of the scene at global level from the input image using five convolutional layers and two fully connected layers. A \ac{ReLU} is used as an activation function after each convolutional layer. To generalise the system, a \ac{LRN} layer was used to perform local normalization. Maxpooling layers were then added after the first and final convolutional layer to reduce the number of parameters. To extract any information from any intermediate layers, the two fully connected layers were added after the final convolutional layer. The gradient network was used to produce two depth gradient maps from the input image. The input image and the predicted depth gradient maps were passed through the refining network to produce a refined depth map. The output from the global depth network, the gradient network and the colour image were used to produce the refined depth map. In their results, \cite{lin2019depth} conclude that this implementation shows that when using the split semantic segmentation task, more relevant information was extracted than when using the depth gradient implementation. Their work was successful in showing that a hybrid architecture of \ac{CNN}s with modular tasks leads to better performance. Additionally, by combining semantic segmentation and depth estimation, performance can be enhanced when compared to tackling the tasks separately. Quantitatively, the results show that a unified hybrid approach outperforms single task approaches with results surpassing other quantitative results for other hybrid approaches \cite{lin2019depth}.

Stereovision has provided very good results for depth estimation in the past years. \cite{scharstein2002taxonomy} provide a detailed evaluation of these stereo systems \cite{scharstein2002taxonomy}. However, stereovision systems are greatly limited by the distance between the two cameras. Depths are worked out using the principle of triangulation between the two captured images. When the distance between the capture point of the images is large, any angle of estimation errors scale upwards. Stereovision is also unsuitable for large patches of objects that have no change in texture. \cite{Saxena2008} propose a system for producing accurate depth maps from two images by rejecting pixels in the images that contain very little texture. They report significant improvements when monocular and stereo cues are used in conjunction to build the combined \ac{MRF} model. Using this combined model, accurate depth maps are produced by leveraging the advantages that both monocular and stereo features offer \cite{Saxena2008}.  The state of the art is \textit{monodepth} \cite{Godard2017}  based on a neural network that learns from a stereovision labelled dataset.

\section{Methodology}

\subsection{Dataset}

The SpatialVOC2K \cite{belz2018spatialvoc2k} dataset is used to train and test the pattern recognition models.  This dataset consists of 2,026 images with object labels, bounding boxes annotations extracted from the \textit{PASCAL VOC2008} challenge dataset \cite{pascal-voc-2008}, to which relations between objects and depth values were added \cite{belz2018spatialvoc2k}. Depth annotation for objects is represented with an integer number between 1 and 100, represent all levels from the foreground to the background as perceived by humans. The set of unique objects in the dataset are ``aeroplane, bird, bicycle, boat, bottle, bus, car, cat, chair, cow, dining table, dog, horse, motorbike, person, potted plant, sheep, sofa, train, TV/monitor'' and there are a total of 9,804 unique object pairs across all images. The images depict both indoor and outdoor scenes. 
For each image object pairs are formed. Object combinations exclude matches between the same object and $n(n - 1)$ pairs of objects are extracted from each image, where $n$ is the number of objects in an image. The images from the SpatialVOC2k \cite{belz2018spatialvoc2k} dataset yielded 11,996 different object combinations which included object and image information. This dataset was filtered to exclude tuples where the object depth was annotated as 'Unspecified' to produce a reduced dataset of 5,244 valid object combinations.

\subsection{Features}

This section gives a brief description as well as some examples of the various features used.  

\textbf{Geometric Features} - These are mostly real-valued  features that are calculated using the object bounding boxes, as in \cite{ramisa-etal-2015-combining,birmingham2019clustering}. These features are normalised with respect to either the image size, the object bounding box or the union bounding box. The geometric features for each object pair were computed using the work previously carried out in \cite{birmingham2019clustering}  to include (1) the area of the objects normalised by the image area and the union area, (2) area of the object overlap, normalised by the minimum area, image area, total area and union area, (3) aspect ratio of both objects, (4) distance between the centroid of the bounding box of the object normalised by image diagonal, union bounding box and union diagonal, (5) distance to size ratio normalized wrt. bounding boxes, (6) euclidian distance between the bounding boxes of the objects normalised by the union area and image area, (7) ratio of the object bounding box limits, (8) ratio of bounding box areas by maximum to minimum ratio and trajector to landmark ratio, (9) trajector centroid position relative to the landmark centroid, (10) trajector position relative to the landmark as a Euclidian vector, (11) trajector centroid position relative to the landmark centroid, vector and unit vector normalised by the union. Some of these features are correlated.  

\textbf{Semantic Language Features} - These are features that are of textual format and are descriptors for the image or for the objects in the image. In other work and approaches, these features are generally encoded using one-hot encoding to convert the textual representation into a vector. The size of the vector depends on the length of the unique samples for the language feature. To work around the issue of data sparseness, \cite{ramisa-etal-2015-combining} employ the use of \textit{word2vec} embeddings \cite{mikolov2013distributed} to encode terms as vectors that represent the term using alternative and related terms while \cite{belz2018spatialvoc2k}  use \ac{GloVe} embeddings \cite{pennington2014glove}.  These are mostly useful for \ac{ZSL} problems where the label of the object is unknown during training since similar objects with similar properties can be related to.

\textbf{Perceptual Features} - These can be described as properties of objects that are identified by the visual system \cite{feldman2014probabilistic}. Studies on humans have shown that perceptual features are task relevant \cite{antonelli2017task}. These features are mostly intuitive. In the PASCAL VOC2008 image dataset, object pose, occlusion, truncation and identification difficulty are examples of features that are of perceptual type \cite{pascal-voc-2008}. 

\textbf{Scene Features} - Scene features offer context to images and the objects contained in them. They are able to extend the bare object features available to further specify the scenario by identifying the scene that best describes the object's location. Incorporating the scene of the image will introduce specificity to the 'scene' at hand \cite{zhou2017places}. \cite{zhou2017places} present \textit{Places-CNNs}, scene classification \ac{CNN}s using state of the art \ac{CNN}s that outperform previous scene classification approaches \cite{zhou2017places}. In this work, the \textit{places-365} scene classifier was used to enhance the image data offered by the selected dataset given the wide scene coverage that is quoted in their work and the large sample of images used for training \cite{zhou2017places}.
A subset of images from the PASCAL Visual Objects Classes 2008 Challenge is used to build an enhanced dataset using relational object information, scene classification information as well as image and object geometrical features \cite{pascal-voc-2008}.
The Places365 scene classification \ac{CNN}
was used to extract the scene categories for the images. 

\subsection{Target Classes}
The objects in the SpatialVOC2k dataset are hand annotated with a depth value ranging from 0 to 100 for each object in the images. The objects depth values were compared to produce the target class labels as  in Table \ref{tab:target_classes_explanation}.  The three-way target classes are \textit{in front of}, \textit{behind} and \textit{neutral}, i.e at similar depth.

\begin{table}[h]
\begin{footnotesize}
\begin{tabular}{c|l}
\textbf{Class} & \textbf{Relation between objects}                                                                       \\ \hline
0              & Object 1 is in front of Object 2                                                                                               \\
\hline
1              & \begin{tabular}[c]{@{}l@{}}Objects are within the tolerance threshold\\
 and are thus considered at equal depth\end{tabular} \\
 \hline
2              & Object 2 is in front of Object 1 \\
\end{tabular}
\end{footnotesize}
\caption{Definitions of the target classes.}
\label{tab:target_classes_explanation}
\end{table}
\cite{Dehaene2011} study the perception of different senses and notions by humans and other animals. In their findings on depth along with other perceived senses, they suggest that humans do not perceive numerical representations of numbers in real world in a proportionate manner but in a logarithmic manner \cite{Dehaene2011}. The human annotations for depth in the \textit{SpatialVOC2K} dataset are bound to introduce a level of noise due to any inconsistencies in the annotations done by the different annotators themselves. Additionally, as \cite{Dehaene2011} have shown, images which have no background, or are captured from an angle which portrays a single object as the background hinder the depth estimation process for the human annotators. To investigate this matter further on the dataset, a number of images from the dataset were selected and the depth annotations of the objects in the images were evaluated. To mitigate the noise from the human annotations, a threshold value was introduced. Four different threshold values were considered for the scope of this experiment; 0, 2, 5 and 10. In this work, the threshold values are considered as the acceptable range where the difference in depth is considered to be within range and thus the objects would be classified as at an equal depth. When varying the threshold, the number of instances per class also changes. This class imbalance would skew the dataset. Higher thresholds will produce a skew with a higher number of neutral classes where the depth is equal (class 1). This will give a high number of neutral true positives and thus increasing the model accuracy. Whilst offering a mitigation for noise in dataset annotation, this threshold must be used with care since high threshold values will introduce skewness in the results. In this work, it is hypothesised that a threshold close or less than 2 needs to be applied as it is the threshold that offers the best balance between the classes for this dataset.

\subsection{Pattern Recognition Models}

\begin{figure}[ht!]
  \centering
  \includegraphics[width=\linewidth]{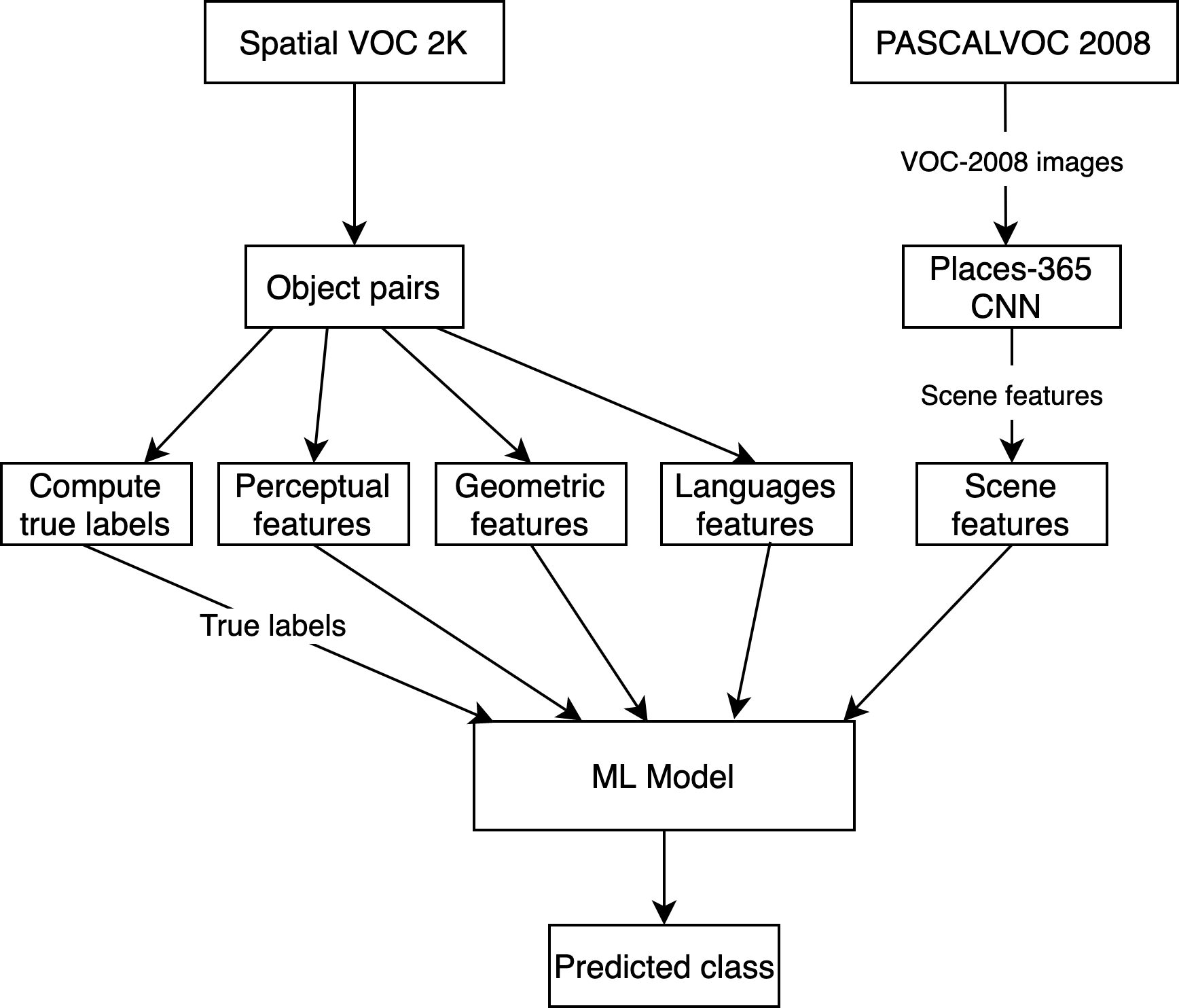}
  \caption{High-level Architecture Diagram.}
  \label{fig:architecture_diagram}
\end{figure}

Figure \ref{fig:architecture_diagram} shows a high-level architecture diagram depicting the dataset feature extraction, target classes and machine learning model.

\begin{figure}[ht!]
  \centering
  \includegraphics[width=1\linewidth]{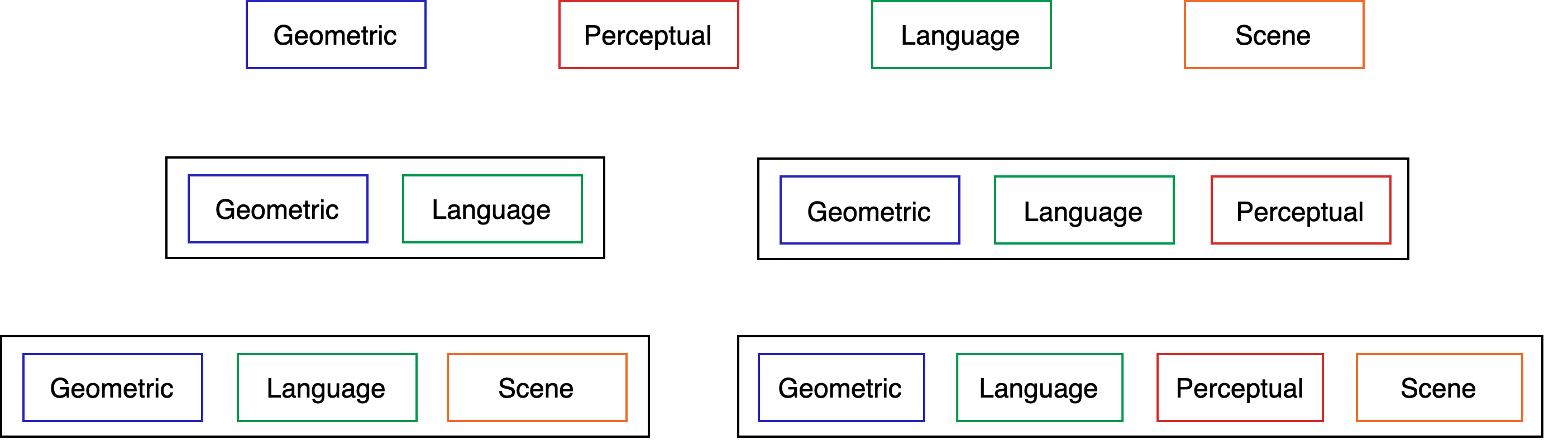}
  \caption[Feature groups and feature groups combinations]{Graphical representation of feature groups and feature group combinations.}
  \label{fig:feature_groups_combinations}
\end{figure}

To avoid training and testing on data that has been co-incidentally split in a manner that favours knowledge of any class, stratified sampling was used. The dataset is split into five folds with each fold containing an equal distribution of the target class. For the experiments, the 51 features extracted were divided into the four feature groups, fig.\ref{fig:architecture_diagram}. The base feature groups are the geometric features, semantic language features, perceptual features and scene features. Four additional feature groups were  formed by concatinating base feature groups. All eight  feature groups, fig.\ref{fig:feature_groups_combinations},  are used in the experiments.

Four different machine learning models are used for the experiments in this work; Decision Tree (DT), Random Forest (RF), Logistic Regressor (LR) and Neural Network (NN). Each of the machine learning models is implemented as a separate script using the Scikit-learn library \cite{scikit-learn}. The features used were normalised and scaled and feature columns with categorical values were one hot encoded. 

An experimental setup was implemented to evaluate and experiment with different models using different feature groups, threshold values and also allow for hyper-parameter tuning. This setup allows for the easy running of several models in bulk with different parameters and configurations. Using the parameters, the possible combinations are extracted and iterated to train, test and evaluate each model. 


\section{Results}

The results are evaluated on the basis of the accuracy metric. A higher accuracy indicates a better model. 
In table \ref{tab:experiment_results}, the two most accurate models for depth thresholds 0 and 2 are shown. Depth thresholds 5 and 10 were used to experiment if higher depth thresholds result in a different accuracy. However, results for depth thresholds higher than 2 are considered to be inaccurate due to large class imbalance that occurs when using such a high tolerance. The experiment results show that the random forest with the geometric, language and perceptual features is the most accurate machine learning model for relative depth prediction when using a depth threshold of 0. Overall, the models using combined feature groups offered better accuracies over using just single feature groups. The accuracy trend of the machine learning models follows this order (descending): random forest, neural network, logistic regressor and decision tree.

\begin{table}
    \centering
    \begin{footnotesize}
    \begin{tabular}{l|l|r|r}
        \textbf{Feature Group} & \textbf{Model} & \textbf{T/H} & \textbf{Accuracy} \\ 
        Geometric \textbf{(Geo)} & RF & 0 & 72.214\% \\ \hline
        Geo & RF & 2 & 71.184\% \\ \hline
        Sem \textbf{(Sem)} & LR & 0 & 53.836\% \\ \hline
        Sem & NN & 0 & 53.702\% \\ \hline
        Perceptual \textbf{(Per)} & NN & 0 & 54.561\% \\ \hline
        Per & RF & 0 & 53.950\% \\ \hline
        Scene & NN & 2 & 43.535\% \\ \hline
        Scene & LR & 2 & 43.077\% \\ \hline
        Geo \& Sem & RF & 0 & 73.130\% \\ \hline
        Geo \& Sem & RF & 2 & 71.069\% \\ \hline
        Geo, Sem \& Per & RF & 0 & \textbf{74.828}\% \\ \hline
        Geo, Sem \& Per & RF & 2 & 72.500\% \\ \hline
        Geo, Sem \& Scene & RF & 0 & 73.073\% \\ \hline
        Geo, Sem \& Scene & RF & 2 & 71.299\% \\ \hline
        Geo, Sem, Per \& Scene & RF & 0 & 74.466\% \\ \hline
        Geo, Sem, Per \& Scene & NN & 0 & 73.340\% \\
        Geo, Sem, Per \& Scene & NN & 2 & \textbf{72.938}\% \\
    \end{tabular}
    \end{footnotesize}
    \caption{Experiment results for depth thresholds 0 and 2 sorted in descending order by accuracy. Model name abbreviations: Random Forest \textbf{(RF)}, Neural Network \textbf{(NN)}, Logistic Regressor \textbf{(LR)} and Decision Tree \textbf{(DT)}.}
    \label{tab:experiment_results}
\end{table}

When analysing the results at depth thresholds 0 and 2 for the top two performing machine learning models, one can notice two salient points; (a) Depth threshold 0 produces a higher accuracy than depth threshold 2; (b) these results are obtained using different feature group sets. For depth threshold 0, the random forest obtains an accuracy of 74.828\% using the geometric, language and perceptual features. On the other hand, at depth threshold 2, the best performing model is the neural network using a combination of all the feature types with a 72.938\% accuracy. Furthermore, the second best performing machine learning model for depth threshold 0 is the neural network and uses all the feature groups. For depth threshold 2, the random forest with the geometric, language and perceptual features achieves the second best model for this threshold with a margin of 0.4\%. The best accuracies for the separate thresholds are very close. For depth threshold 0, they differ by 1.489\% whilst for depth threshold 2 the best performing model differs by 0.438\%. Individually, the geometrical features are most effective at predicting depth and the accuracies are close to the best accuracies, whilst the scene features are the least useful.
For the rest of the evaluation, the best performing models for the random forest and neural network at depth thresholds 0 and 2 will be considered.


\section{Evaluation}

In \cite{birmingham2018adding} machine generated depth features are used to study their usefulness when predicting prepositions that describe the spatial relationship between two objects depicted in an image. To obtain the pixel level depth the authors in \cite{birmingham2018adding} make use of the 
monodepth model \cite{Godard2017}
and process the depthmap produced by the model to obtain average depth values for the objects inside the bounding boxes. 


To  compare the results achieved to those of the \textit{monodepth} depth prediction \ac{CNN}, the images from the SpatialVOC2K were passed through the \textit{monodepth} \ac{CNN} to obtain the pixel level depth values of each image. The depth map was then segmented using the object bounding boxes.  Following \cite{birmingham2018adding}, two different object depth values were computed using these pixels; (a) \textit{Average depth per object} - The average of the pixels found in the enclosing bounding box; (b) \textit{Radially weighted depth per object} - When extracting the average depth, the distance of the pixel from the centroid of the object is taken as a weight. A greater importance is given to the predicted depth value for the pixels that are closer to the center of the object whereas further pixels which may be incorrectly labelled due to edge errors are given less importance.  The evaluation is subdivided in two parts to evaluate the \textit{monodepth} \ac{CNN} agreement with the SpatialVOC2K dataset and to compare the best performing relative depth estimation models produced with the depth produced by the \textit{monodepth} \ac{CNN}.



With reference to comparing \textit{monodepth} derived relative depths to SpatialVOC2k human annotations derived relative depths, table \ref{tab:monodepth_agreement_with_human_annotations} gives the accuracy scores obtained for the average and \ac{RWA} results across the four thresholds. These results show that for threshold 0, the \textit{monodepth} \ac{CNN} and the SpatialVOC2K dataset are $\approx$ 60.65\% in agreement. When using the average depth, the \ac{CNN} is in a higher agreement level than when using the \ac{RWA} since for all instances, the average depth accuracy is higher than the \ac{RWA} accuracy. When using depth threshold 1, inferior results can be noted when compared to depth threshold 0. The best accuracy over all the accuracies can be reported for threshold 0.

\begin{table}[h]\centering
\begin{tabular}{@{}r|rrr@{}}
\textbf{Threshold} & 0 & 1 & 5 
\\
\textbf{Average} & 0.6065 & 0.5885 & 0.5208 
\\
\textbf{\ac{RWA}} & 0.5981 & 0.5824 & 0.5059 
\\
\end{tabular}
\caption[Accuracy score comparison between the \textit{monodepth} \ac{CNN} depths and the \textit{SpatialVOC2k} depth results at different thresholds]{Accuracy score comparison between the \textit{monodepth} \ac{CNN} depths and the \textit{SpatialVOC2k} depth results at different thresholds. Accuracy score is between 0 and 1. }
\label{tab:monodepth_agreement_with_human_annotations}
\end{table}


The percenatge agreement between the best performing models identified in this work (i.e. the random forest and the neural network) for depth thresholds 0 and 2 and the \textit{monodeph} predictor are computed.
Table \ref{tab:dissertation_models_comparison_CNN} shows the results obtained for the random forest and neural network at depth thresholds 0 and 2. 
The RF model agrees $\approx$ 67\% with the \textit{monodepth}  predictor. On the other hand, the \textit{monodepth} predictor agrees 60\% of the time with the human annotations.

\begin{table}
\begin{footnotesize}
\begin{tabular}{ccccc}

T/H & Model  & Accuracy & Mono-Mean & Mono-RWA\\
\hline
0 & RF & 74.82\% & 67.50\% & 67.20\% \\
0 & NN & 73.40\% & 55.62\% & 54.96\% \\
	\hline
2 & RF & 72.50\% & 55.63\% & 54.96\% \\
2& NN & 72.90\% & 52.53\% & 52.73\% \\
\end{tabular}
\end{footnotesize}
\caption{Comparison of model agreement with the \textit{monodepth} depth prediction \ac{CNN} for the RF and NN at depth thresholds 0 and 2.}
\label{tab:dissertation_models_comparison_CNN}
\end{table}

The model agreement with the \textit{monodepth} \ac{CNN} shows that the relative depth prediction model (the random forest) and the \ac{CNN} seem to struggle to predict the correct relative depth for the same object pair instances. This confirms the previous hypothesis with regards to the noise level in the dataset. Nonetheless, the random forest using the geometric, language and perceptual features with the depth threshold set to 0 produces a relative depth prediction model which is 74.82\% accurate. This accuracy surpasses the 60.65\% accuracy that the \textit{monodepth} depth prediction \ac{CNN} was able to achieve by an accuracy boost of 14.17\% in relative depth prediction.

\section{Conclusion and Future Work}

A relative depth estimation model is implemented from geometric, language and perceptual features  and for some tasks it may be more effective than more complex models resulting in lower computational costs.  A maximum of 74.828\% accuracy is achieved with a random forest model, which translates to a 14\% increase in accuracy when compared to results obtained from the \textit{monodepth} neural network model. 
  The next step is to study its use in vision and language tasks.

\bibliographystyle{apalike}
{\small
\bibliography{sample-base}}

%

\end{document}